# A Bayesian Interpretation of the Particle Swarm Optimization and Its Kernel Extension

**Peter Andras***

School of Computing Science, Newcastle University, Newcastle upon Tyne, United Kingdom

## Abstract

Particle swarm optimization is a popular method for solving difficult optimization problems. There have been attempts to formulate the method in formal probabilistic or stochastic terms (e.g. bare bones particle swarm) with the aim to achieve more generality and explain the practical behavior of the method. Here we present a Bayesian interpretation of the particle swarm optimization. This interpretation provides a formal framework for incorporation of prior knowledge about the problem that is being solved. Furthermore, it also allows to extend the particle optimization method through the use of kernel functions that represent the intermediary transformation of the data into a different space where the optimization problem is expected to be easier to be resolved–such transformation can be seen as a form of prior knowledge about the nature of the optimization problem. We derive from the general Bayesian formulation the commonly used particle swarm methods as particular cases.

**Funding:** There was no specific funding provided for this research beyond the support from the School of Computing Science of Newcastle University. The funders had no role in study design, data collection and analysis, decision to publish, or preparation of the manuscript.

**Competing Interests:** The author has declared that no competing interests exist.

* E-mail: peter.andras@ncl.ac.uk

## Introduction

Particle swarm optimization (PSO) is a heuristic optimization method that was proposed in the mid-1990s following inspiration from social problem solving (e.g. collaborative foraging by flocking birds) [1]. The key idea of the method is to combine individual and social learning in an optimization context, such that the social component can speed up the individual search for the optimal solution of a problem. The PSO algorithm performs a parallel search for the optimal solution with many individual searches associated with particles, and collaborative influencing of these by the joint best performance of all searches [2–5]. As a result the particles move as a swarm in the space of problem solutions, searching for the best solution. The expectation is that due to the parallel search for the optimal solution and the social swarming of the separate parallel searches, the likelihood of finding the optimal or a near-optimal solution is high.

PSO has been applied to a range of problems, including optimization in the context of management of power systems (e.g. optimal distribution, reliability management) [6], [7], scheduling of operations [8], [9], vehicle routing and loading optimization [10], scheduling of applications on computer grids [11], and estimation of parameters in complex industrial systems [12]. In general, it is considered to be a good choice for practical solution of optimization problems when the problem is high-dimensional; is defined by multiple criteria and potentially conflicting constraints; or it is of complex combinatorial nature [3].

PSO in general is applied to optimization problems where potential solutions are defined as vectors, i.e. vectors of solution parameters ([5]). For each potential solution there is a characteristic performance value that defines the goodness of the solution. Each particle has an associated solution parameter vector and moves in the space of these vectors with the aim of finding the optimal solution of the problem. The optimal solution has the highest characteristic performance value. Each particle has an associated velocity vector that is updated in each optimization turn and is used to update the solution parameter vector associated with the particle. The common equations driving the particle swarm optimization are the following ([3]):

$$v_i(t+1) = w \cdot v_i(t) + \varphi \cdot r_\varphi \cdot (x_i^b - x_i(t)) + \eta \cdot r_\eta \cdot (x^g - x_i(t)) \quad (1)$$

$$x_i(t+1) = x_i(t) + v_i(t+1) \quad (2)$$

where $x_i(0)$ and $v_i(0)$are the original solution parameter and velocity vectors, $x_i^b$and $x^g$are the solution parameters found so far that are the best among those found by particle $i$ and the best among those found by all particles, $w$ is the inertial factor, $\varphi$ and $\eta$ are attraction parameters of the optimization process and $r_\varphi$ and $r_\eta$ are random numbers drawn from the uniform distribution over (0,1).

Another common variant of the equations was proposed around 2000 [13], [14] with the aim to improve the convergence and avoid the divergence of the search paths of particles. This variant uses a constricted version of equation (1):

$$v_i(t+1) = \chi \cdot \left( w \cdot v_i(t) + \varphi \cdot r_\varphi \cdot (x_i^b - x_i(t)) + \eta \cdot r_\eta \cdot (x^g - x_i(t)) \right) \quad (3)$$

$$\chi = \frac{2}{\left|2-\varphi-\eta-\sqrt{(\varphi+\eta)^2-4(\varphi+\eta)}\right|} \quad (4)$$

where it is assumed that $\varphi+\eta>4$. An alternative version is to just use $\chi$ instead of $w$ in equation (1) [13]. There have been proposed several other variants of the equations (1) and (3) on the basis of various heuristics and also combinations of these with other methods forming hybrid PSO algorithms (see reviews of these variants in [2–5]).

There have been attempts to introduce a less heuristic and more formal theoretical foundation for the PSO [15–18]. These approaches (e.g. bare bones PSO [15], [16]) propose to replace equations (1) and (2) with an appropriate choice of the next particle position driven by the sampling of a distribution that is calculated given the individual and global optimal solutions found so far ($x_i^b$ and $x^g$). Others [17], [18] suggested to use a Kalman filter calculation to estimate the distribution that is sampled for the generation of the next position of the particles. (See Section 2 for further discussion.)

Here we propose a new theoretical approach to PSO starting from a Bayesian perspective. We describe a generalized version of PSO in terms of Bayesian optimization of prospective solutions of a problem. This allows the progressive incorporation into the algorithm of information about the problem that is discovered as the PSO proceeds and also the incorporation of prior knowledge about the nature of possible problem solutions. We show how this generalized Bayesian PSO leads to the particular cases that correspond to the original PSO, the bare bones PSO and the Kalman filter based PSO. Using the generalized Bayesian PSO formulation we also extend the PSO to consider prior knowledge information about the problem domain using kernel functions and leading to a nonlinear version of the Bayesian PSO. We also demonstrate the performance of the proposed Bayesian variants of the PSO algorithm using a set of commonly used test functions [19].

## Related Works

The PSO algorithm simulates the movement of a swarm of agents searching for an optimal position on a landscape defined by a problem and its possible solutions. The inspiration comes from the physical movement of animals (e.g. ants) on this landscape. However, it has been realized that the constraints that apply to physical movements on the landscape according to the original inspiration are not strictly necessary for the realization of the optimization process by the PSO. Kennedy [15] proposed to replace the landscape constrained movement by a 'flying' movement that ignores such constraints and allows free movement in a probabilistic sense according to the available information about the possible problem solutions.

The resulting bare bones PSO [15], as its name suggests, aims to use the minimal necessary components of the PSO algorithm to achieve its optimization objectives. The modified algorithm is based on the realization that the positions of the particles follow probability distributions defined by the individual and global best positions found so far; i.e., $x_i^b$ and $x^g$. According to the original bare bones PSO the next position of particle $i$ is picked randomly from the normal distribution with the vector mean value $\mu_i = \frac{1}{2}(x_i^b + x^g)$ and having the covariance matrix $C_i$ defined by the elements $C_i^{jk}=0$, if $j \neq k$, and $c_i^{jj} = |x_{i,j}^b - x_j^g|$. An alternative choice for the covariance matrix is $C_i = \frac{1}{2}||x_{i,j}^b - x_j^g|| \cdot I_m$, where $I_m$ is the $m \times m$ identity matrix, and $m$ is the dimensionality of the solution vectors [20]. Thus the bare bones PSO eliminates the need of consideration of the velocity of particles and focuses on movement of the particles driven by the calculated probability distributions that are sampled to determine the next position of the particles.

A theoretical advantage of the bare bones PSO is that it replaces to some extent the heuristic inspiration of the standard (original) PSO algorithm by more clear theoretical foundations through the sampling of probability distributions for the selection of positional updates of the particles. The choice of the normal distribution is driven partly by the observation of the data from standard PSO and partly by convenience [15], [20].

Further work on the bare bones PSO led to the use of alternative choices for the distributions that sampled for the generation of the next positions of the particles [16], [20], [21]. Such alternatives are the use of appropriate Levy or Cauchy distributions. These distributions may fit better certain problems solved by PSO. It should be noted that the choice of the distribution from which the next position is picked for the particles is determined heuristically on the basis of knowledge of the nature of the problem that is aimed to be solved by the PSO.

Another similar approach was proposed by Monson and Seppi [17], [18] who suggested using the formalism of Kalman filters to calculate the distribution from which the next position of the particles is chosen. The underlying idea is that the movement of particles in the swarm can be considered in terms of processes where the underlying state representing the closeness to the true optimal solution is hidden, and the observable data are the current particle position and the global optimal position. In this conceptualization the underlying state can be estimated using a Kalman filter for each particle. In effect the Kalman filter for each particle determines a normal distribution from which the next value of position of the particle can be sampled.

The Kalman filter PSO first defines the following vectors $y_i(t) = (x_i(t), x_i(t) - x_i(t-1))^T$, $z(t) = (x^g(t), 0)^T$, $q_i(t) = (x_i^b(t), 0)^T$, and $\overline{y_i}(0) = y_i(0)$. Using these vectors the equations of the Kalman filter PSO for each particle $i$ are as follows:

$$K_i(t) = (FW_i(t)F^T + W_y)H^T\left(H(FW_i(t)F^T + W_y)H^T + W_z)^{-1}\right) \quad (5)$$

$$\overline{y_i}(t+1) = F\overline{y_i}(t) + K_i(t)((I_{2m} - Q)q_i(t) + Qz(t) - HF\overline{y_i}(t)) \quad (6)$$

$$W_i(t) = (I_{2m} - K_i(t))(FW_i(t)F^T + W_y) \quad (7)$$

where $W_i(0)$, $W_y$ and $W_z$ are covariance matrices that characterize the noise in the measurement of $y_r(0)$, and $y$ and $z$ in general, $Q$ is a matrix representing the balance between the influence of the global and particle specific optima (i.e. corresponds to $\lambda_b$ and $\lambda_g$), $I_{2m}$ is the $2m \times 2m$ identity matrix, and $0_m$ is the $m \times m$ zero matrix. The algorithm also has specific assumptions about the $F, H, Q$ matrices, e.g. $H = (I_m, 0_m)$.

The value of $x_i(t+1)$, i.e. the next position of the particle $i$, is given by taking a sample $\hat{y}_i(t+1) = (x_i(t+1), \hat{v}_i(t+1))^T$ from the normal distribution with mean vector $F\overline{y_i}(t+1)$ and covariance matrix $W_i(t)$. The covariance matrices $W_i(0), W_y$ and $W_z$ are set heuristically.

Thus the Kalman filter PSO provides an alternative to the bare bones PSO for finding a distribution for each particle from which the next position of the particle can be determined by sampling. An issue of the Kalman filter PSO is that the velocity appears nominally in the sample $\hat{y}_i(t+1)$, but this part of the vector is ignored. The theory of the algorithm does not provide a clear explanation for this. The specific setting of the $F,H,Q$ matrices also constrains very much the generality of the algorithm. This algorithm, just as the bare bones PSO, relies on some heuristic assumptions, in this case about the covariance matrices $W_i(0), W_y$ and $W_z$ [18].

## Bayesian Interpretation

The optimization problem that we try to solve using PSO can be formulated in general by considering that the characteristic performance value corresponding to a possible solution is defined by an unknown function $f(x)$ (a fitness function of the solution) where $x$ is the solution parameter vector [7]. It is assumed that values of $f(x)$ can be evaluated for any $x$, but this is an 'expensive' operation – here evaluation of the unknown function means that there is an available experimental test that returns the value of this function, or an approximation of this; consider for example oil exploration, where the function describing the distribution of oil quantity underground is not known, but by an expensive drilling and small-scale extraction operation can be estimated at any point of the oil exploration area. It is also assumed that $\frac{\partial}{\partial x} f(x)$ cannot be evaluated directly and to calculate its numerical approximation is prohibitively 'expensive'. The aim of the optimization problem is to find $x^*$ such that

$$x^* = \arg\max_{x \in X} f(x) \tag{8}$$

Where $X$ is the domain of solution parameter vectors within which we search for the optimal solution. Possibly, the evaluation of $f(x)$ for any given $x$ may be noisy, i.e. the evaluation of the potential solution parameter vector $f(x)$ gives as result the value $\hat{f}(x) = f(x) + \xi$, where $\xi$ is a random value that follows a noise distribution – typically a Gaussian distribution centered at 0, i.e. $N(0,\sigma)$. To keep things simple, we assume that $\hat{f}(x) > 0$ for all $x \in X$ - this does not reduce the generality of the analysis since for any function we can add a constant to make its values positive everywhere, assuming that these values are bounded for the considered domain of solution parameter vectors.

### General Formulation

At the beginning of the search for the optimal solution we might have some prior knowledge about the likelihood of possible parameter vectors to be the optimal solution. Alternatively we might have some general belief about these likelihoods, for example if $X \subseteq \mathbf{R}^m$ contains an m-dimensional ball centered at 0 with radius equal to 1, then we may start with a default assumption that the likelihood of being the optimal solution is given by a Gaussian probability density function centered at the 0 vector and with a covariance matrix given by the $m \times m$ identity matrix. In general, the prior knowledge or the default assumption is given in the form of a probability density function $P(x)$ defined over $X$. For the sake of simplicity, in continuation we assume that $X = \mathbf{R}^m$, where $m$ is the dimensionality of the solution parameter vectors.

Let us assume that we start the search with $n$ particles which have their associated solution parameter vectors $x_i(0)$ (i.e. the position vectors of the particles). Evaluating the candidate solutions we find the performance value estimates $\hat{f}(x_i(0))$. This information is used to update our beliefs about the likelihood of a given parameter vector $x$ being the optimal solution of the optimization problem. Using the Bayes theorem we can write

$$P\left(x \middle| \hat{f}(x_i(0), i=1,\ldots,n\right) = \frac{P\left(\hat{f}(x_i(0), i=1,\ldots,n \middle| x\right) \cdot P(x)}{P\left(\hat{f}(x_i(0), i=1,\ldots,n\right)} \tag{9}$$

The denominator on the right-hand side is common for all $x$, so it can be ignored as part of the normalizing constant of the distribution. $P\left(\hat{f}(x_i(0), i=1,\ldots,n|x\right)$ is the likelihood of finding the values $\hat{f}(x_i(0), i=1,\ldots,n$, if $x$ is assumed to be the optimal solution. This is the same as the likelihood of finding $x$ as the optimal solution if $\hat{f}(x_i(0), i=1,\ldots,n$ are given, and the prior assumption is that all $x$ have equal probability to be the optimal solution. To calculate this likelihood we may assume that the likelihood of $x_i(0)$ being the optimal solution is proportional to $\hat{f}(x_i(0)$ and that it also depends on the function evaluation values for all other considered $x_j(0)$ – we call this the dependence assumption. Then we construct an approximation of the corresponding probability density function as a linear combination of basis functions anchored at $x_i(0)$, i.e

$$P\left(\hat{f}(x_i(0)), i=1,\ldots,n \middle| x\right) = \frac{1}{\sum_{i=1}^{n} \hat{f}(x_i(0))} \sum_{i=1}^{n} \hat{f}(x_i(0)) \cdot G(x; x_i(0)) \tag{10}$$

where $G(x; x_i(0))$ are such that $\int_{\mathbf{R}^m} G(x; x_i(0)) dx = 1$, we also assume that $\sum_{i=1}^{n} \hat{f}(x_i(0)) > 0$. For example, a common choice is to set $G(x; x_i(0)) = \exp\left(-\frac{\beta}{2}(x - x_i(0))^T (x - x_i(0))\right)$, i.e. a Gaussian distribution centered at $x_i(0)$ with a scaled identity matrix being the covariance matrix. Note that if the optimization problem is a minimization problem, i.e. finding $\arg\min_{x \in X} f(x)$ is the objective, then the assumption is that the likelihood of $x_i(0)$ being the optimal solution is proportional to $1/\hat{f}(x_i(0))$ and in equation (10) $\hat{f}(x_i(0))$ are replaced by $1/\hat{f}(x_i(0))$.

Alternatively, we may assume that $x_i(0)$ being the optimal solution is independent of any other $x_i(0)$ being the optimal solution, and that the certainty of $x_i(0)$ being the optimal solution depends on $\hat{f}(x_i(0))$ – we call this the independence assumption. In this case joint probability distribution for likelihood of finding $x$ as the optimal solution if $\hat{f}(x_i(0), i=1,\ldots,n$ are given can be written as

$$P\left(\hat{f}(x_i(0)), i=1,\ldots,n \middle| x\right) = \left(\prod_{i=1}^{n} G(x; x_i(0))^{\hat{f}(x_i(0))}\right)^{\frac{1}{\sum_{i=1}^{n} \hat{f}(x_i(0))}} \tag{11}$$

with similar assumptions as above, in the case of equation (10). In the cases when the optimization problem is a minimization problem the same argument applies as above and in equation (11) $\hat{f}(x_i(0))$ are replaced by $1/\hat{f}(x_i(0))$.

To improve our estimate of the optimal solution parameter vector the PSO method selects the next set of position vectors in the space of the solution parameter vectors. The bare bones PSO [15] introduced the idea that the next set of vectors can be chosen by sampling distributions over the space of these vectors, such that the distributions are defined by the latest best estimates of the solution. The Kalman filter PSO [17] also proposed a similar approach to calculate a distribution for each particle, which is then sampled to generate the next vector associated with the particle. We follow this idea in a general sense, by sampling the posterior $P\left(x|\hat{f}(x_i(0)),i=1,\ldots,n\right)$ distribution to get the parameter vectors $x_i(1)$, i.e. the probability of picking $x_i(1)$ is given by $P\left(x_i(1)|\hat{f}(x_i(0)),i=1,\ldots,n\right)$.

To keep the notations consistent and meaningful we denote

$$P_0(x)=P(x) \tag{12}$$

and the posterior distribution following the evaluation of the $t$-th sample of parameter vectors is denoted by

$$P_{t+1}(x)=\frac{P\left(\hat{f}(x_i(t)),i=1,\ldots,n|x\right)\cdot P_t(x)}{P\left(\hat{f}(x_i(t)),i=1,\ldots,n\right)} \tag{13}$$

This means that the sample vectors $x_i(t+1)$ are generated by sampling the distribution $P_{t+1}(x)$. By generating consecutive samples of possible solution vectors the posterior distribution gets increasingly constrained, and this is likely to lead increasingly closer to the actual optimal solution.

The calculation of $P_{t+1}(x)$ following the dependence assumption leads to the formula

$$P_{t+1}(x)=a_t\cdot P\left(\hat{f}(x_i(t)),i=1,\ldots,n|x\right)\cdot P_t(x)= \\ =\left(\prod_{j=1}^{t} a_j\right)\cdot P_0(x)\cdot\prod_{j=1}^{t}\frac{1}{\sum_{i=1}^{n}\hat{f}(x_i(t))}\sum_{i=1}^{n}\hat{f}(x_i(t))\cdot G(x;x_i(j)) \tag{14}$$

where $a_t$ are normalizing constants – i.e. to make the integral of the distribution equal to one over the whole definition domain.

The alternative independence assumption leads to the following formula:

$$P_{t+1}(x)=a_t\cdot P\left(\hat{f}(x_i(t)),i=1,\ldots,n|x\right)\cdot P_t(x)= \\ =\left(\prod_{j=1}^{t} a_j\right)\cdot P_0(x)\cdot\prod_{j=1}^{t}\left(\prod_{i=1}^{n}\left(G(x;x_i(j))^{\hat{f}(x_i(t))}\right)\right)^{\frac{1}{\sum_{i=1}^{n}\hat{f}(x_i(t))}} \tag{15}$$

## Bayesian PSO with Gradient Ascent Optimization

An alternative to the generation of a sample from the distribution given in equation (14) or (15) is to generate the next set of position vectors by updating of the previous set of position vectors. In principle the best estimate of the optimal solution parameter vector, given the posterior $P_{t+1}(x)$, is the vector $x^*_{t+1}$ for which $P_{t+1}(x)$ reaches its maximum. Thus we could try to find $x^*_{t+1}$ by solving the equation

$$\frac{\partial}{\partial x}P_{t+1}(x)=0 \tag{16}$$

Unfortunately, in principle this equation cannot be solved. Thus, taking the last set of position of vectors $x_i(t),i=1,\ldots,n$, we could calculate the next set of position vectors $x_i(t+1),i=1,\ldots,n$ by using gradient ascent updates of the current vectors. The formula of $P_{t+1}(x)$ contains a large product that would make difficult the calculation of gradient ascent updates as suggested above. Instead we can apply the gradient ascent updates by considering the optimization of $\ln P_{t+1}(x)$. In this case, following the dependence assumption we have

$$\ln P_{t+1}(x)= \\ \sum_{j=1}^{t}\ln a_j+\ln P_0(x)+\sum_{j=1}^{t}\ln\left(\frac{1}{\sum_{i=1}^{n}\hat{f}(x_i(j))}\sum_{i=1}^{n}\hat{f}(x_i(j))\cdot G(x;x_i(j))\right) \tag{17}$$

and

$$\frac{\partial}{\partial x}\ln P_{t+1}(x)= \\ \frac{\frac{\partial}{\partial x}P_0(x)}{P_0(x)}+\sum_{j=1}^{t}\frac{1}{\sum_{i=1}^{n}\hat{f}(x_i(j))\cdot G(x;x_i(j))}\sum_{i=1}^{n}\hat{f}(x_i(j))\cdot\frac{\partial}{\partial x}G(x;x_i(j)) \tag{18}$$

If we follow the independence assumption we get

$$\ln P_{t+1}(x)= \\ \sum_{j=1}^{t}\ln a_j+\ln P_0(x)+\sum_{j=1}^{t}\frac{1}{\sum_{i=1}^{n}\hat{f}(x_i(j))}\sum_{i=1}^{n}\hat{f}(x_i(j))\cdot\ln G(x;x_i(j)) \tag{19}$$

and

$$\frac{\partial}{\partial x}\ln P_{t+1}(x)= \\ \frac{\frac{\partial}{\partial x}P_0(x)}{P_0(x)}+\sum_{j=1}^{t}\frac{1}{\sum_{i=1}^{n}\hat{f}(x_i(j))}\cdot\left(\sum_{i=1}^{n}\hat{f}(x_i(j))\cdot\frac{1}{G(x;x_i(j))}\cdot\frac{\partial}{\partial x}G(x;x_i(j))\right) \tag{20}$$

Thus, we can calculate the gradient updates and the formula for the next set of sample vectors is

$$x_i(t+1)=x_i(t)+\gamma\cdot\frac{\partial}{\partial x}\ln P_{t+1}(x)\Big|_{x_i(t)} \tag{21}$$

where $\gamma$ is the learning constant of the gradient ascent updates.

This approach may work more efficiently than the sampling of the full posterior distribution although it is more constrained in terms of the actual sampling (i.e. the new position vectors are constrained by the current position vectors). The price that we pay for this additional constraint is that we may sample more frequently than it would be implied by the full posterior distribution, regions of the space where the likelihood of finding the optimum is low, and conversely we may sample less frequently other regions, where the full posterior distribution would indicate high likelihood of finding the optimal solution.

## Summary

In principle the advantage of the Bayesian approach is that we take into account the full distribution of likelihoods of finding the optimal solution at any admissible solution parameter vector. Of course, the disadvantage comes in computational terms since maintaining and updating the information about this full distribution is computationally expensive.

The Bayesian approach to PSO described here provides a general insight into how PSO algorithms work in principle. It shows that the PSO algorithm makes the position vectors converge towards the most likely location of the solution vector. This convergence is improved stepwise as the increasing amount of data constraints more and more the estimated distribution of the likely location of the solution vector. This distribution converges towards the actual distribution of the solution vector(s) that may be a Dirac δ distribution, if there is a single solution, or combination of Dirac δ and possibly uniform distributions if there are multiple solutions (note that the uniform distribution is the case if there is a part of the solution vector space over which the optimized function takes the same value (i.e. it is constant), which is the optimal value). In the next section we consider particular cases to show the link between the Bayesian interpretation of PSO and the practically used PSO methods.

# Particular Cases

## Gaussian PSO

First, we introduce the Gaussian PSO, which is a particular version of the Bayesian interpretation of the PSO. We choose Gaussian distributions for the distributions involved in our calculations,

$$P_0(x)=\alpha_0 \exp\left(-\frac{1}{2}||x||^2\right) \tag{22}$$

and

$$G(x;x_i(j))=\alpha_i^j \exp\left(-\frac{\beta}{2}||x-x_i(j)||^2\right) \tag{23}$$

Where $\alpha_0$ and $\alpha_i^j$ are appropriate constants to satisfy the integral requirement of the distributions. Following the dependence assumption we find the update formulas

$$\frac{\partial}{\partial x}\ln P_{t+1}(x)\Big|_{x_r(t)}= -x_r(t)-\sum_{j=1}^{t}\frac{1}{\sum_{i=1}^{n}\hat{f}(x_i(j))\alpha_i^j \exp\left(-\frac{\beta}{2}||x_r(t)-x_i(j)||^2\right)} \cdot\left(\sum_{i=1}^{n}\hat{f}(x_i(j))\cdot\alpha_i^j \exp\left(-\frac{\beta}{2}||x_r(t)-x_i(j)||^2\right)\cdot\beta\cdot(x_r(t)-x_i(j))\right) \tag{24}$$

and

$$x_r(t+1)= (1-\gamma)\cdot x_r(t)+\gamma\cdot\sum_{j=1}^{t}\frac{1}{\sum_{i=1}^{n}\hat{f}(x_i(j))\alpha_i^j \exp\left(-\frac{\beta}{2}||x_r(t)-x_i(j)||^2\right)} \cdot\left(\sum_{i=1}^{n}\hat{f}(x_i(j))\cdot\alpha_i^j \exp\left(-\frac{\beta}{2}||x_r(t)-x_i(j)||^2\right)\cdot\beta\cdot(x_r(t)-x_i(j))\right) \tag{25}$$

Alternatively if we follow the independence assumption we get the update formulas

$$\frac{\partial}{\partial x}\ln P_{t+1}(x)\Big|_{x_r(t)}= -x_r(t)-\sum_{j=1}^{t}\frac{1}{\sum_{i=1}^{n}\hat{f}(x_i(j))}\cdot\left(\sum_{i=1}^{n}\hat{f}(x_i(j))\cdot\beta\cdot(x_r(t)-x_i(j))\right) \tag{26}$$

and

$$x_r(t+1)= (1-\gamma)\cdot x_r(t)+\gamma\cdot\sum_{j=1}^{t}\frac{1}{\sum_{i=1}^{n}\hat{f}(x_i(j))}\cdot\left(\sum_{i=1}^{n}\hat{f}(x_i(j))\cdot\beta\cdot(x_r(t)-x_i(j))\right) \tag{27}$$

We note that the $(1-\gamma)$ multiplicative factor disappears if $P_0(x)$ is assumed to be uniform distribution. The meaning of this factor is that if the second additive term on the right side of the equations (25) and (27) is zero then the position vectors converge to the zero vector that is consistent with the assumption of equation (22). If the $P_0(x)$ is a uniform distribution and the second additive term of the equations (25) and (27) is zero, the position vectors do not move, again this being consistent with the assumption of the uniform initial prior distribution.

The equations (25) and (27) are similar to the position vector update equations used in the standard PSO algorithm (equations (1) and (2)) with the exception of the random components. The random components are replaced by the multipliers that depend on the evaluations of the function $f$. Our equations are derived in a principled manner from the consideration of the posterior distribution of likelihoods of possible solution parameter vectors being the optimal solution vector. The similarity between the equations confirms the correctness of the intuition that led to the formulation of PSO algorithms, and explains the success of application of heuristic PSO algorithms (i.e. these algorithms move the position vectors towards the likely solution vectors given the available data).

## Standard PSO

Let us start by making the choice of distributions further more specific. We modify the functions $G(x;x_i(j))$ such that we take into account only a selection of these corresponding to certain $x_i(j)$ vectors. Let us first define

$$\varepsilon_i^j = \begin{cases} 1, \text{ if } x_i(j)=x_i^b(j) \text{ and } x_i^b(j)\neq x_i^b(j-1) \text{ or } x_i(j)=x^g(j) \text{ and } x^g(j)\neq x^g(j-1) \\ 0, \text{ otherwise} \end{cases} \quad (28)$$

where

$$x_i^b(j)=\underset{k=1,\dots,j}{\arg\max}\hat{f}(x_i(k)) \quad (29)$$

$$x^g(j)=\underset{\substack{k=1,\dots,j\\ i=1,\dots,n}}{\arg\max}\hat{f}(x_i(k)) \quad (30)$$

Then we may define the modified versions of these functions as follows:

$$\tilde{G}(x;x_i(j))=\begin{cases} G(x;x_i(j)), \text{ if } \varepsilon_i^j=1 \\ c, \text{ otherwise} \end{cases} \quad (31)$$

Where $c$ is a constant. If we follow the dependence assumption we set $c=0$. Alternatively, if we follow the independence assumption we set $c>0$, defining a uniform distribution over a bounded region in the space of solution parameter vectors where we search for the optimal solution.

Considering the Gaussian distributions introduced in equations (22) and (23) together with the modifications as defined in equations (28) – (31), after calculations we find that the revised update equations for solution parameter vectors associated with particles are as follows for the two kinds of joint distribution assumptions (i.e. dependence and independence):

$$x_r(t+1)=(1-\gamma)\cdot x_r(t)+\gamma\cdot\sum_{j=1}^{t}\frac{1}{\sum_{i=1}^{n}\varepsilon_i^j\cdot\hat{f}(x_i(j))\alpha_i^j\exp\left(-\frac{\beta}{2}||x_r(t)-x_i(j)||^2\right)} \cdot\left(\sum_{i=1}^{n}\varepsilon_i^j\cdot\hat{f}(x_i(j))\cdot\alpha_i^j\exp\left(-\frac{\beta}{2}||x_r(t)-x_i(j)||^2\right)\cdot\beta\cdot(x_r(t)-x_i(j))\right) \quad (32)$$

and

$$x_r(t+1)= (1-\gamma)\cdot x_r(t)+\gamma\cdot\sum_{j=1}^{t}\frac{1}{\sum_{i=1}^{n}\varepsilon_i^j\cdot\hat{f}(x_i(j))}\cdot\left(\sum_{i=1}^{n}\varepsilon_i^j\cdot\hat{f}(x_i(j))\cdot\beta\cdot(x_r(t)-x_i(j))\right) \quad (33)$$

noting that if $\varepsilon_i^j=0$ for all $i$ for a given $j$ then the $j$-th additive term disappears (i.e. otherwise we would have division by 0).

Next, we constrain the set of distributions considered for each particle to those distributions that are associated with best vectors calculated for this particle and those associated with temporary global maxima $x^g(j)$. In formal terms this is implemented using a 0 multiplier for functions associated with other particles in the case of the calculation of posterior distribution according to the dependence assumption, and by using a 0 exponent for distributions associated with other particles in the case of the posterior distribution calculation following the independence assumption, with the exception that this change does not apply if the vector associated with another particle is the temporary global optimum vector. This means that for any $j$, at most only two of the vectors $x_i(j), i=1,\dots,n$, are considered, i.e. if $x_r(j)=x_r^b(j)$ or $x_r(j)=x^g(j)$ then $x_r(j)$ is considered, or if $x_i(j)=x^g(j)$ then $x_i(j)$ is considered, and the $j$-th additive term is present in the sum, otherwise it disappears.

Now, let us assume that the information that we gained earlier in the process of optimization is discounted as time passes and as we expect to get closer to the true optimal solution. To implement the discounting of information we modify the distributions that we considered by changing the value of the parameter â for distributions associated with earlier positions of the particles. First, we define

$$\theta_i^{j,k}=\begin{cases} 1, \text{ if } x_i(j)=x_i^b(k) \text{ and } x_i(j)=x^g(k) \\ 0, \text{ otherwise} \end{cases} \quad (34)$$

We make the distributions adaptive, by setting

$$G^k(x;x_i(j))=\begin{cases} G^{k-1}(x;x_i(j)), \text{ if } \theta_i^{j,k}=1 \\ \alpha_i^j\cdot\exp\left(-\frac{\beta\cdot\tau^{k-j}}{2}||x-x_i(j)||^2\right), \text{ otherwise} \end{cases} \quad (35)$$

Where $0<\tau<1$. This means that with every turn of updating the solution parameter vectors associated with particles the distributions associated with earlier positions of the particles get flatter as $\beta\cdot\tau^{k-j}$ approaches 0.

Replacing $\beta$ in equation (32) and (33) by $\beta\cdot\tau^{k-j}$ means that additive terms corresponding to earlier particle-specific and global temporary optima get discounted. We can rewrite equations (32) and (33) in a common form as

$$x_r(t+1)= (1-r)\cdot x_r(t)+r\cdot\left(\beta_t^g\cdot(x^g(t)-x_r(t))+\beta_t^b\cdot\left(x_r^b(t)-x_r(t)\right)\right)+r\cdot\tau\cdot v_r(t) \quad (36)$$

Where $\beta_t^g$ and $\beta_t^b$ are calculated according to (32) or (33) and $v_r(t)$ is the sum of the additive terms corresponding to earlier temporary global and particle specific optima.

Equation (36) is the same as the update equation of the standard PSO with the exception that the random factors are replaced by $\beta_t^g$ and $\beta_t^b$, which vary according to the evaluations of the optimized function. The speed momentum component is represented by $y\cdot\tau\cdot v_r(t)$. If $\tau$ is set sufficiently small then this latter component will decrease quickly and can be considered vanishingly small. This variant of equation (36) is equivalent to the original PSO equation that did not include the speed momentum term.

## Bare Bones PSO

The bare bones PSO introduced by Kennedy [15] has been described briefly in Section 2. This variant of PSO generates the next vectors associated with particle $r$ by sampling the Gaussian distribution centered at $\frac{1}{2}\left(x_r^b(j)+x^g(j)\right)$ and having a covariance matrix defined as $\frac{1}{2}||x_r^b(j)-x^g(j)||\cdot I_m$, where $I_m$ is the $m\times m$ identity matrix.

In our Bayesian approach this is equivalent of fully discounting the prior information ($P_0(x)$) and also the past prior to the last evaluation of the solution parameter vectors associated with the particles. Furthermore, the bare bones PSO assumes that the posterior distribution considered for particle $r$ in equation (13) depends only on $x_r^b(j)$ and $x^g(j)$. Thus, this distribution is:

$$P_r^{bb}(x|\hat{f}(x_i(t)), i=1,\ldots,n) = \frac{1}{(2\pi \cdot ||x_r^b(j)-x^g(j)||)^{\frac{m}{2}}} \cdot \exp\left(-\frac{1}{2\cdot||x_r^b(j)-x^g(j)||}\cdot\left\|x-\frac{1}{2}(x_r^b(j)+x^g(j))\right\|^2\right) \quad (37)$$

Further calculations of this distribution lead to the following formula

$$\begin{aligned} P_r^{bb}(x|\hat{f}(x_i(t)), i=1,\ldots,n) = \\ \frac{1}{(2\pi \cdot ||x_r^b(j)-x^g(j)||)^{\frac{m}{2}}} \cdot \exp\left(-\frac{1}{2\cdot||x_r^b(j)-x^g(j)||}\cdot||x-x_r^b(j))||^2\right) \\ \cdot \frac{1}{(2\pi \cdot ||x_r^b(j)-x^g(j)||)^{\frac{m}{2}}} \cdot \exp\left(-\frac{1}{2\cdot||x_r^b(j)-x^g(j)||}\cdot||x-x^g(j))||^2\right) \\ \cdot (2\pi \cdot ||x_r^b(j)-x^g(j)||)^{\frac{m}{2}} \cdot \exp\left(-\frac{||x_r^b(j)-x^g(j)||}{8}\right) \end{aligned} \quad (38)$$

This distribution is the same as the distribution corresponding to equation (26) in the context of Gaussian PSO with the independence assumption for the posterior by setting $\beta = ||x_r^b(j)-x^g(j)||^{-1}$, and assuming that $x_r^b(j) \neq x^g(j)$. If $x_r^b(j) = x^g(j)$ the posterior becomes a Dirac-δ distribution centered at $x_r^b(j) = x^g(j)$. Thus, the Bayesian interpretation shows that the bare bones PSO is equivalent of the standard PSO with a specific setting of $\beta$.

## Kalman Filter PSO

Let us start by considering the posterior distribution defined following the independence assumption (equation (15)) and let us assume the following component distributions

$$G(x; x_i(j)) = \alpha_i^j \exp\left(-\frac{1}{2}(x-x_i(j))^T V_i(j)(x-x_i(j))\right) \quad (39)$$

and define $\tilde{G}(x; x_i(j))$ according to equation (31).

This means that the posterior for particle $r$ is a product of Gaussian distribution, which itself is a Gaussian distribution

$$P_{r,t}(x) = c_{r,t} \exp\left(-\frac{1}{2}(x-\overline{x_r}(t))^T \overline{V_r}(t)(x-\overline{x_r}(t))\right) \quad (40)$$

Considering that.

$$\begin{aligned} P_{r,t+1}(x) = \alpha_r^g \exp\left(-\frac{\lambda_g}{2}(x-x^g(t))^T V_r^g(t)(x-x^g(t))\right) \\ \cdot \alpha_r^b \exp\left(-\frac{\lambda_b}{2}(x-x_r^b(t))^T V_r^b(t)(x-x_r^b(t))\right) \cdot P_{r,t}(x)^{1-\lambda_g-\lambda_b} \end{aligned} \quad (41)$$

Where $\lambda_g$ and $\lambda_b$ are determined by $\hat{f}(x_i(j))$ according to equation (15), we can write the update equations for $\overline{x_r}(t)$ and $\overline{V_r}(t)$. After some calculations we find

$$\overline{V_r}(t+1) = (1-\lambda_g-\lambda_b)\overline{V_r}(t) + \lambda_g V_r^g(t) + \lambda_b V_r^b(t) \quad (42)$$

$$\begin{aligned} \overline{x_r}(t+1) = \\ \overline{V_r}(t+1)^{-1}\left((1-\lambda_g-\lambda_b)\overline{V_r}(t)\overline{x_r}(t) + \lambda_g V_r^g(t)x^g(t) + \lambda_b V_r^b(t)x_r^b(t)\right) \end{aligned} \quad (43)$$

An alternative way to calculate this covariance matrix and mean vector is to build a Kalman filter following the way proposed in the Kalman filter PSO algorithm [17]. This algorithm applies the Kalman filter calculations to combined position and velocity vectors for each prototype. The velocity is calculated according to

$$v_r(t+1) = \lambda_b(x_r^b(t)-x_r(t)) + \lambda_g(x^g(t)-x_r(t)) \quad (44)$$

Let us consider the vectors $y_r(t) = (x_r(t), v_r(t))^T$, $z_r(t) = (x^g(t), 0)^T$, and $q_r(t) = (x_r^b(t), 0)^T$ and the Kalman filter PSO equations (5) – (7). We consider $Q = \begin{pmatrix} Q_x & 0_m \\ 0_m & 0_m \end{pmatrix}$ the matrix representing the balance between the influence of the global and particle specific optima (i.e. corresponds to $\lambda_b$and $\lambda_g$), and

$$F = \begin{pmatrix} I_m & I_m \\ 0_m & I_m \end{pmatrix} \quad (45)$$

$$H = (I_m \quad 0_m) \quad (46)$$

$$W_r(t) = \begin{pmatrix} V_r^{x,x}(t) & V_r^{x,v}(t) \\ V_r^{x,v}(t)^T & V_r^{v,v}(t) \end{pmatrix} \quad (47)$$

$$W_y = \begin{pmatrix} V_{x,x} & V_{x,v} \\ V_{x,v}^T & V_{v,v} \end{pmatrix} \quad (48)$$

$$W_z = \begin{pmatrix} V_{g,g} & 0_m \\ 0_m & 0_m \end{pmatrix} \quad (49)$$

and $I_m$ is the $m \times m$ identity matrix, $0_m$ is the $m \times m$ nil matrix, $V_r^{x,x}(t) = \overline{V_r}(t)$, $V_r^{x,v}(t)$, and $V_r^{v,v}(t)$ are the covariance matrices for $x$, $x$ and $v$, and $v$ for particle $r$ at time $t$, and $V_{x,x}, V_{x,v}, V_{v,v}$, and $V_{g,g}$ are fixed component matrices of $W_y$ and $W_z$.

Considering the specific forms of the matrices given in equations (45) – (49), and the equations (42) – (43), we define

$$K_r(t+1) = \begin{pmatrix} K_r^x(t+1) \\ K_r^v(t+1) \end{pmatrix} \quad (50)$$

and identify $V_r^{x,x}(t) = \overline{V_r}(t)$. Assuming that $\overline{v_r}(0) = 0$, implying that the original mean velocity for any particle $r$ is zero (the covariance matrix of velocity vectors is given by the $V_{v,v}$ component of $W_y$), after some calculations we find

$$\overline{V_r}(t+1)=\overline{V_r}(t)+V_{x,x}-(\overline{V_r}(t)+V_{x,x})(\overline{V_r}(t)+V_{g,g})^{-1}(\overline{V_r}(t)+V_{x,x}) \quad (51)$$

$$\begin{aligned}&\overline{x_r}(t+1)= \\ &\left(I_m-(\overline{V_r}(t)+V_{g,g})^{-1}(\overline{V_r}(t)+V_{x,x})\right)\overline{x_r}(t)+ \\ &(\overline{V_r}(t)+V_{g,g})^{-1}(\overline{V_r}(t)+V_{x,x})Q_x x^g(t)+ \\ &(\overline{V_r}(t)+V_{g,g})^{-1}(\overline{V_r}(t)+V_{x,x})(I_m-Q_x)x_r^b(t)\end{aligned} \quad (52)$$

The last two equations are equivalent of a reformulation of equations (42) and (43) by considering the setting of $\lambda_g$, $\lambda_b$, $V_r^g(t)$ and $V_r^b(t)$ such that

$$\begin{aligned}&\overline{V_r}(t+1)\left(I_m-(\overline{V_r}(t)+V_{g,g})^{-1}(\overline{V_r}(t)+V_{x,x})\right)= \\ &(1-\lambda_g-\lambda_b)\overline{V_r}(t)\end{aligned} \quad (53)$$

$$\overline{V_r}(t+1)(\overline{V_r}(t)+V_{g,g})^{-1}(\overline{V_r}(t)+V_{x,x})Q_x=\lambda_g V_r^g(t) \quad (54)$$

$$\overline{V_r}(t+1)(\overline{V_r}(t)+V_{g,g})^{-1}(\overline{V_r}(t)+V_{x,x})(I_m-Q_x)=\lambda_b V_r^b(t) \quad (55)$$

Thus, the Kalman filter PSO can be seen as a particular case of the Gaussian PSO with the independence assumption.

### Other Particular Cases

We may fully discount the past, and consider all current vectors associated with particles to determine the updates of the vectors. The Bayesian formulation of the PSO allows the calculation of the weights that apply to the vectors associated with all particles. Depending on whether we follow the dependence or independence assumptions, we get the following update equations in the case of the algorithm variants based on Gaussian PSO:

$$\begin{aligned}&x_r(t+1)=(1-\gamma)\cdot x_r(t)+ \\ &\gamma\cdot\beta\cdot\sum_{i=1}^{n}\frac{\hat{f}(x_i(t))\alpha_i^t\exp\left(-\frac{\beta}{2}||x_r(t)-x_i(t)||^2\right)}{\sum_{i=1}^{n}\hat{f}(x_i(t))\alpha_i^t\exp\left(-\frac{\beta}{2}||x_r(t)-x_i(t)||^2\right)}\cdot(x_i(t)-x_r(t))\end{aligned} \quad (56)$$

and

$$x_r(t+1)=(1-\gamma)\cdot x_r(t)+\gamma\cdot\beta\cdot\sum_{i=1}^{n}\frac{\hat{f}(x_i(t))}{\sum_{i=1}^{n}\hat{f}(x_i(t))}\cdot(x_i(t)-x_r(t)) \quad (57)$$

The difference between the two formulations is that the dependence assumption implies the use of weightings that depend on the relative positions of these vectors, while such weights are not used in the case of the independence assumption.

The potential advantage of the consideration of the full current set of vectors associated with particles for the calculation of the next vectors is that in certain cases the selection of the global best and particle best vectors may impose an undesirable bias that may temporarily steer the particles away from the actual optimum. The weighted impact of the vectors allows a more balanced calculation of the vector updates, reducing the chance of such undesirable bias.

Another alternative to reduce this kind of bias is to consider the full set of past global best vectors for each update, in combination with temporal discounting. This is implemented in the context of Bayesian PSO in a similar manner that we used to derive the Bayesian interpretation of the standard PSO (see equations (28) – (36)). The update equation for solution parameter vectors in this case is

$$x_r(t+1)=(1-\gamma)\cdot x_r(t)+\gamma\cdot\beta\cdot\sum_{j=1}^{t}\tau^{t-j}\cdot(x^g(j)-x_r(t)) \quad (58)$$

## Kernel Extension

As we noted in Section 3 the Bayesian approach allows to incorporate prior knowledge about the optimization problem into the application of the PSO method. In some cases such prior knowledge may be expressed in the sense that certain distributional assumptions are valid about the likelihood of a vector being the optimal solution, given the evaluation values associated with other vectors. For example, Gaussian distributions of likelihood of being the optimal vector may be valid in a transformed space defined by some transformation of the original solution parameter vectors. Generally, such prior knowledge can be considered as regularization constraints [22] that are known to apply to the unknown function for which the location of the optimal value is searched for.

In continuation we assume that such prior knowledge implies that combinations of Gaussian distributions in a transformed space are valid indicators of the likelihood of a vector being the optimal vector, given the knowledge of the evaluation of current vectors associated with particles (either in the summative or multiplicative sense – i.e. assuming dependence or independence of probabilities implied by evaluation of different vectors associated with particles). The transformation of the space is denoted by $\Psi$. Furthermore, we assume that $\Psi$ is such that the internal product in the transformed space can be expressed using a kernel function defined over pairs of vectors of the original space of solution parameter vectors, i.e. the transformation $\Psi$ corresponds to a Mercer kernel $K$ [23]:

$$<\Psi(x),\Psi(y)>=K(x,y) \quad (59)$$

Following this assumption, we find that

$$||\Psi(x)-\Psi(y)||^2=K(x,x)+K(y,y)-2K(x,y) \quad (60)$$

Thus, equations (22) and (23) can be rewritten as follows

$$P_0(x)=\alpha_0\exp\left(-\frac{1}{2}(K(x,x)+K(0,0)-2K(x,0))\right) \quad (61)$$

and

$$G(x;x_i(j))= \alpha_i^j \exp\left(-\frac{\beta}{2}(K(x,x)+K(x_i(j),x_i(j))-2K(x,x_i(j)))\right) \quad (62)$$

Many valid kernel functions are defined such that $K(x,y)=H(||x-y||)$. Considering such kernels implies that $K(x,x)=c$, where $c$ is a constant (often $c=0$ or $c=1$). Following further calculations along the lines of equations (17) – (21) we find the update equations for vectors associated with particles. In the case of the dependence assumption the update equation becomes

$$x_r(t+1)=x_r(t)-\gamma\frac{\partial}{\partial x}K(x,0)\Big|_{x_r(t)}+ \gamma\cdot\sum_{j=1}^{t}\left(\sum_{i=1}^{n}\hat{f}(x_i(j))\cdot\alpha_i^j\exp(-\beta(c-K(x_r(t),x_i(j)))\right)^{-1} \cdot\left(\sum_{i=1}^{n}\hat{f}(x_i(j))\cdot\alpha_i^j\exp(-\beta(c-K(x_r(t),x_i(j)))\cdot\beta\cdot\frac{\partial}{\partial x}K(x,x_i(j))\Big|_{x_r(t)}\right) \quad (63)$$

In the case of the independence assumption we find

$$x_r(t+1)=x_r(t)-\gamma\frac{\partial}{\partial x}K(x,0)\Big|_{x_r(t)}+ \gamma\cdot\sum_{j=1}^{t}\left(\sum_{i=1}^{n}\hat{f}(x_i(j))\right)^{-1}\cdot\left(\sum_{i=1}^{n}\hat{f}(x_i(j))\cdot\beta\cdot\frac{\partial}{\partial x}K(x,x_i(j))\Big|_{x_r(t)}\right) \quad (64)$$

There are many options for kernel functions, for example [23], [24]:

$$K(x,y)=\sqrt{||x-y||^2+\mu} \quad (65)$$

$$K(x,y)=\frac{\mu}{||x-y||}\sin\frac{||x-y||}{\mu} \quad (66)$$

$$K(x,y)=\frac{1-\frac{1}{2}\cos(x-y)}{\frac{5}{4}-\cos(x-y)} \quad (67)$$

$$K(x,y)=\cos(\sin(x))\cdot e^{\cos(x)} \quad (68)$$

where $\mu$ is a strictly positive constant.

The particular cases of Bayesian PSO are naturally adapted to the kernel extension of PSO. The vector update equation for the kernel version of the standard PSO becomes

$$x_r(t+1)=x_r(t)+ r\cdot\left(\beta_t^g\cdot\frac{\partial}{\partial x}K(x,x^g(t))\Big|_{x_r(t)}+\beta_t^b\cdot\frac{\partial}{\partial x}K\left(x,x_r^b(t)\right)\Big|_{x_r(t)}\right)+r\cdot\tau\cdot v_r(t) \quad (69)$$

Considering the kernel function given in equation (68) the kernel extension of the standard PSO update equations is

$$x_r(t+1)=x_r(t)+r\cdot\left(\beta_t^g\cdot\sin\left(\|x_r(t)-x^g(t)\|^2+\sin\left(\|x_r(t)-x^g(t)\|^2\right)\right)\cdot e^{\cos\left(\|x_r(t)-x^g(t)\|^2\right)}\right) \cdot(x_r(t)-x^g(t))+y\cdot\left(\beta_t^b\cdot\sin\left(\|x_r(t)-x_r^b(t)\|^2+\sin\left(\|x_r(t)-x_r^b(t)\|^2\right)\right)\cdot e^{\cos\left(\|x_r(t)-x_r^b(t)\|^2\right)}\right) \cdot\left(x_r(t)-x_r^b(t)\right)+r\cdot\tau\cdot v_r(t) \quad (70)$$

The practical meaning of this variant of the PSO vector update equation is that the best approach towards the optimum given the global best and particle best vectors becomes variable according to the trigonometric functions involved in the equation (69). In a sense the role of the variability of particle position vector updates represented by the use of trigonometric functions replaces the variability induced by the random numbers in equation (1). In this case this is done in a principled manner since the kernel function (equation (68)) is assumed to represent some knowledge about the nature of the optimization problem.

## Application Examples

To evaluate the performance of Bayesian PSO we compared the standard PSO (equations (1) and (2) ); the bare bones PSO (see Section 2); two kinds of Gaussian PSO: representing the dependence and independence assumption versions of Gaussian PSO, corresponding to equations (25) and (27) – Gaussian 1 and Gaussian 2, respectively; and a kernel extension of the standard PSO (equation (70) with $\tau=0$). To compare the performance of these methods we chose the following 10 dimensional functions [19]:

1) Axis-parallel hyper-ellipsoid: $\mathbf{u}\in[-100,100]^{10}$

$$f(\mathbf{u})=\sum_{k=1}^{10}(ku_k)^2 \quad (71)$$

2) Griewank: $\mathbf{u}\in[-600,600]^{10}$

$$f(\mathbf{u})=\frac{1}{4000}\sum_{k=1}^{10}u_k^2-\prod_{k=1}^{10}\cos\left(\frac{u_k}{\sqrt{k}}\right)+1 \quad (72)$$

3) Rastrigin: $\mathbf{u}\in[-5.12,5.12]^{10}$

$$f(\mathbf{u})=100+\sum_{k=1}^{10}(u_k^2-10\cos(2\pi u_k)) \quad (73)$$

4) Rosenbrock: $\mathbf{u}\in[-30,30]^{10}$

$$f(\mathbf{u})=\sum_{k=1}^{9}(100(u_{k+1}-u_k^2)^2+(u_k-1)^2) \quad (74)$$

5) Salomon: $\mathbf{u}\in[-100,100]^{10}$

$$f(\mathbf{u}) = 1 - \cos\left(2\pi\sqrt{\sum_{k=1}^{10} u_k^2}\right) + 0.1\sqrt{\sum_{k=1}^{10} u_k^2} \quad (75)$$

6) Schwefel: $\mathbf{u} \in [-500,500]^{10}$

$$f(\mathbf{u}) = 5000 + \sum_{k=1}^{10} -u_k \sin(\sqrt{|u_k|}) \quad (76)$$

7) Sphere: $\mathbf{u} \in [-100,100]^{10}$

$$f(\mathbf{u}) = \sum_{k=1}^{10} u_k^2 \quad (77)$$

8) Step: $\mathbf{u} \in [-5.12,5.12]^{10}$

$$f(\mathbf{u}) = 60 + \sum_{k=1}^{10} \lfloor u_k \rfloor \quad (78)$$

9) Modulus sum: $\mathbf{u} \in [-5.12,5.12]^{10}$

$$f(\mathbf{u}) = 60 + \sum_{k=1}^{10} |u_k| \quad (79)$$

For each function we ran 100 times each algorithm with random initialization of 100 particles. The aim of the optimization in all cases is to find the minimum value of the function. The stop condition of the runs was either reaching 100,000 iterations or reduction of the variability of the particle positions around the global best position to be close to zero, i.e

$$\frac{1}{10m}\sum_{r=1}^{m} ||x_r - x^g||^2 < 0.001 \quad (80)$$

The performance of a method for a given function is characterized by the mean value of the minimum values for the given function that were found by the respective PSO method and the standard deviation of the mean minimum values. To compare the performances of different methods for each considered optimized function we used the two-tailed t-test. The mean convergence times of the PSO methods varied, the comparably fastest being the bare bones and the Gaussian PSOs, the standard PSO was somewhat slower than these (needing more iteration steps), and the kernel standard PSO was the slowest, needing many more iterations than the other methods.

The generic forms of some of the algorithms were slightly modified to facilitate their execution and testing. In the case of bare bones PSO we sampled only the middle of the distribution, i.e. instead of $C_i = \frac{1}{2}||x_i^b - x^g|| \cdot I_m$ we used $C_i' = \frac{1}{5}C_i$ as the covariance matrix – this improved very much the results making them more comparable with the results of the other PSO variants.

For the Gaussian PSO algorithms we retained only the last 100 positions for each particle to calculate the estimated probability density function of the optimal solution. The parameters that we chose for the Gaussian PSOs and the kernel standard PSO were: $\gamma = 0.8$ for all, $\beta = 0.4$ for the dependence assumption Gaussian PSO and the kernel standard PSO, and $\beta = 0.1$ for the independence assumption Gaussian PSO.

The performance results are presented in Table 1. Table 2 presents the statistical comparison of these using the t-test. The results show that the bare bones PSO is statistically significantly better than the standard PSO for all functions with the exception of the Rosenbrock and modulus sum functions. They also show that the Gaussian PSOs are statistically very significantly better than the bare bones PSO for seven out nine functions, the exceptions being the Schwefel and step functions. The results show that the kernel standard PSO is significantly better than the bare bones PSO for all functions except the Rastrigin function.

## Discussion and Conclusions

The Bayesian interpretation of PSO provides a principled basis for the analysis of PSO algorithms. We have shown in this paper that special cases of the Gaussian PSO variant of the Bayesian PSO are equivalent of the standard PSO [1], bare bones PSO [15] and Kalman filter PSO [17]. The Bayesian interpretation of PSO allows formal analysis of the mechanisms and performance factors of PSO algorithms and this can lead to a better understanding of the reasons why certain PSO algorithms may work better in certain circumstances than other similar algorithms.

In general, we have to assume that PSO algorithms are applied to optimization problems that are complex and the evaluation of the optimized function is costly. This means that extensive sampling and evaluation of the optimized function is not feasible and thus it is impractical to approximate this function or to approximate numerically the derivatives of the function. The implication of this is that searching for the optimal solution parameter vector cannot be driven by usual optimization methods that can be applied to find optima of analytically tractable known functions. Thus a possible and feasible alternative is that the search for the optimal solution parameter vector (or an approximation of this) is driven by estimating the probability distribution of the location of the optimal solution given the available data about the evaluations of the optimized function at positions corresponding to particles of the PSO algorithm. This approach is represented by the Bayesian interpretation of PSO.

The dimensionality of the argument vectors of the optimized function (i.e. the position vectors of the particles) and the number of particles in the PSO algorithm have significant impact on the effectiveness of the PSO algorithm. In order to estimate the density function of the probability distribution of the likely position of the optimal solution vector we need a sufficient sample of the space of the position vectors. As the PSO algorithm proceeds we gain additional sample data and the approximation of the true probability distribution gets improved. Note that the true distribution might be a Dirac δ distribution centered at unique the global optimum of the function, if this exists. Alternatively it is possible that the true distribution is a linear combination of Dirac δ distributions centered at equivalent global optima of the function. It is also possible that this distribution is an extension of the Dirac δ distribution and/or uniform distribution in cases when the optimum positions form together a surface or a sub-space in the space of the argument vectors of the optimized

**Table 1.** Performance Results of the PSO Algorithms.

| Function | Standard | Bare Bones | Gaussian 1 | Gaussian 2 | Kernel Standard |
|---|---|---|---|---|---|
| Hyper-ellipsoid | 286778.3 (158032) | 230254.4 (148784) | 753.12 (31.05) | 87.294 (4.853) | 8446.3 (11112) |
| Griewank | 118.026 (47.05) | 93.129 (50.12) | 0.8712 (0.009) | 0.9128 (0.010) | 2.5708 (1.925) |
| Rastrigin | 77.298 (21.96) | 64.846 (22.62) | 49.043 (0.856) | 2.0539 (0.101) | 82.04 (25.81) |
| Rosenbrock | 5.318E+9 (6E+10) | 4.708E+9 (5E+10) | 3.376E+7(3.3E+6) | 2786.52 (375.51) | 2019.62 (3378) |
| Salomon | 11.763 (2.82) | 10.446 (3.12) | 0.6842 (0.011) | 0.2336 (0.005) | 2.6078 (1.389) |
| Schwefel | 3079.66 (244.822) | 2620.42 (290.559) | 3479.845 (28.51) | 3466.609 (24.893) | 1660.44 (456.97) |
| Sphere | 14925.6 (6837.9) | 9137.47 (554.59) | 29.053 (0.938)) | 3.4561 (0.183) | 42.543 (7.709) |
| Step | 31.14 (3.7578) | 27.24 (3.2694) | 29.96 (0.488) | 32.06 (0.3575) | 4.56 (5.6126) |
| Modulus sum | 14.296 (3.6839) | 13.212 (4.2015) | 2.8209 (0.051) | 0.23 (0.005) | 7.141 (3.774) |

Mean Value (Standard Deviation).

function (e.g. the step function in Section 6). If the number of particles is sufficiently large we can choose from a wide range of possible component distributions ($G(x;x_i(j))$)in equations (14) and (15)). However, if the particles sample the space of the solution parameter vectors very sparsely the reasonable choices for component distributions are reduced to distributions that represent the simplest distributional assumptions (e.g. Gaussian, exponential, uniform distributions).

In the introduction of the Bayesian PSO algorithms we used the assumption that in the case of an optimization problem that is formulated as a maximization of a positive function, the likelihood of $x_i$ being the optimal solution is proportional to $\hat{f}(x_i)$ and in the case of the dependence assumption it also depends on the function evaluation values for all other considered $x_j$. In the case of minimization problems we replaced $\hat{f}(x_i)$ by $1/\hat{f}(x_i)$. These are the simplest assumptions. However, in principle it could be assumed that in the case of a maximization problem the likelihood of $x_i$ being the optimal solution is proportional to $\mu(\hat{f}(x_i))$, where $\mu(x)$ is monotonous positive function that represents prior knowledge about the relationship between $\hat{f}(x_i)$ and the likelihood of $x_i$ being the optimal solution. Similarly, in the case of a minimization problem we could use $1/\mu(\hat{f}(x_i))$ instead of $1/\hat{f}(x_i)$ if prior knowledge indicates the appropriateness of $\mu(x)$. Furthermore, the requirement of $\mu(x)$ being a monotonous positive function may also be relaxed if the prior knowledge about the problem is sufficient to make the assumption of a non-monotonous $\mu(x)$ appropriate for the problem. For example, if in the case of a maximization problem it is known that the interesting maxima are above 100, then we may assume that the use of a $\mu(x)$ such that $\mu(x)=0$ if $x<100$ is appropriate. Of course, the use of $\mu(x)$ can be incorporated into the component probability density functions $G(x;x_i)$ that are used for the calculation of the posterior distributions in equations (14) and (15).

The proposed Bayesian PSO in principle takes into account the full information gathered through the use of algorithm about the nature of the optimization problem that is being solved. In practice the range of this information may be cut in order to increase the speed of the algorithm, as we have shown in the case of the particular variants of the PSO (see Section 4). The Bayesian interpretation of the PSO provides a principled way of incorporating any part of the additional information that may not be considered by the usual variants of the algorithm (i.e. the information provided by the evaluation of parameter vectors by the particles as they pass through the parameter space). The Bayesian PSO also allows to incorporate into the algorithm prior information about the optimization problem that is being solved. This prior information may simply be represented by a prior distribution over the problem space that indicates likely locations of the optimal solution (i.e. $P_0(x)$), or it may be expressed through the use of an appropriate kernel function used through the kernel extension of the Bayesian PSO (see Section 5).

**Table 2.** Comparison of Performance Results of the PSO Algorithms.

| Function | Bare Bones vs Standard | Gaussian 1 vs Bare Bones | Gaussian 2 vs Bare Bones | Kernel Standard vs Bare Bones |
|---|---|---|---|---|
| Hyper-ellipsoid | 0.01003* | 4.30E-28* | 3.51E-28* | 4.25E-27* |
| Griewank | 0.00037* | 9.85E-34* | 1.01E-33* | 3.9E-33* |
| Rastrigin | 0.0001* | 1.41E-9* | 1.39E-48* | 1.21E-6* |
| Rosenbrock | 0.44226 | 2.16E-15* | 1.54E-15* | 1.54E-15* |
| Salomon | 0.00205* | 3.37E-53* | 6.41E-55* | 9.45E-49* |
| Schwefel | 2.16E-25* | 1.43E-52* | 7.34E-55* | 2.68E-40* |
| Sphere | 4.6E-10* | 5.02E-30* | 4.09E-30* | 5.57E-30* |
| Step | 3.09E-13* | 7.27E-6* | 3.83E-19* | 1.75E-21* |
| Modulus sum | 0.14762 | 6.32E-134* | 1.05E-143* | 1.75E-21* |

t-test p-values (* indicates significance, below 0.05 p-value).

The use of the kernel version of PSO algorithms replaces the random variation inducing elements of the PSO algorithm with a similar, but more principled, source of variation, which is provided by the inclusion of the kernel function into the equations (see equation (69)). The kernel function represents prior knowledge about the nature of the optimized function. In principle, this allows to improve the effectiveness of the PSO algorithm even if the number of particles is relatively small in comparison with the dimensionality of the space of argument vectors of the optimized function. This is because the use of the kernel function is expected to implicitly drive the search along appropriate lower dimensional surfaces within the high dimensional space, thus improving the effective sampling of argument vector space (i.e. the more effective sampling is with respect the lower dimensional surface on which the search proceeds).

If the function that is optimized is very variable the Bayesian PSO may need finely tuned parameters ($\gamma$ and $\beta$) to achieve good results. An alternative way to improve its performance is to sample the distribution specified in equation (14) or (15) just as in the case of the bare bones PSO, instead of using a variant of the equation (21) to generate a deterministic update of the particle position vectors. Sampling of the distributions will make the algorithm computationally more expensive, but at the same time it allows more faithful guidance towards the actual optimum position than the deterministic updating.

The Bayesian interpretation of PSO algorithms paves the way for many future developments in PSO research. By providing solid theoretical foundations for the analysis of PSO algorithms and their performance factors it is expected to stimulate the work on variants of PSO and hybrids of PSO with other computational methods. In particular, the research about the choice of appropriate kernels and component distributions that represent prior knowledge about the optimization problem to be solved is likely to attract attention. This is because such appropriate choices can make very significant differences in the performance of the PSO algorithm and this might expand considerably the areas of effective practical applications of PSO algorithms.

**Note.** The Delphi code of the Gaussian PSO algorithms and of the kernel PSO algorithm discussed in this paper are available on request from the author. To request a copy of the algorithm codes please email to peter.andras@ncl.ac.uk.

## Author Contributions

Conceived and designed the experiments: PA. Performed the experiments: PA. Analyzed the data: PA. Contributed reagents/materials/analysis tools: PA. Wrote the paper: PA.